\documentclass[cameraready]{Interspeech}
\title{Towards a Phonology-Informed Evaluation of Multilingual TTS}

\author[affiliation={1}, orcid=0009-0002-0073-1113,correspondingauthor]{Sneha}{Ray Barman}
\author[affiliation={1,2}, orcid=0000-0002-6391-7532]{Neeraj Kumar}{Sharma}
\author[affiliation={1,3}, orcid=0000-0002-4254-9080]{Shakuntala}{Mahanta}
\address{
    $^1$ Centre for Linguistic Science \& Technology, IIT Guwahati \\
    $^2$ Mehta Family School for Data Science \& Artificial Intelligence, IIT Guwahati  \\
    $^3$ Department of Humanities \& Social Sciences, IIT Guwahati
}

\email{[sneha.barman, neerajs, smahanta]@iitg.ac.in}


\usepackage{comment}
\usepackage{hyperref}
\usepackage{multirow}
\usepackage{tipa}
\usepackage{adjustbox}
\usepackage{xcolor}
\usepackage{subcaption} 

\begin{document}

\maketitle

% the abstract here must exactly match the abstract entered into the paper submission system
\begin{abstract}
    % 1000 characters. ASCII characters only. No citations.
Neural TTS systems can sound natural across languages, but naturalness does not guarantee the preservation of sound contrasts that distinguish words from their grammatical forms. Standard metrics like MOS do not test for this. We propose a classifier-based framework that audits TTS output against language-specific phonological patterns using human speech as a benchmark. Testing Assamese advanced tongue root (ATR) vowel harmony with Meta's MMS TTS, we show that a classifier trained on human speech transfers to synthesized speech with minimal loss. The faithfulness audit reveals that [+ATR] mid vowels are realized as [-ATR] in 1/3 tokens despite an underlying [+ATR] specification, a bias absent in human speech. At the word level, predicted ATR labels classify harmony more accurately than transcription labels, indicating a gap between intended and produced phonology. The framework offers task-specific diagnostics and generalizes to other phonological contrasts with measurable acoustic cues.
%naturalness does not guarantee that 
%out-of-fold
\end{abstract}

\section{Background}

Neural architectures and the massive scaling of training data have rapidly improved multilingual text-to-speech (TTS) systems. Standard evaluation metrics based on Mean Opinion Scores (MOS) focus on perceived naturalness and whether words are recoverable by automatic recognition. However, sounding natural does not guarantee that a system reproduces a language's sound patterns, especially when grammatical context determines which sounds appear together. \cite{malisz2019modern} showed that modern neural TTS has largely closed the perceived naturalness gap with human speech. This makes the problem more pressing: if a system scores well on MOS, systematic phonological errors may go entirely undetected.
%Standard metrics focus on how natural something sounds and whether words can be recognized, but not whether the synthesized speech is phonologically consistent with the language it represents.  
%\cite{malisz2019modern} highlighted how modern neural TTS has largely closed the perceived naturalness gap with human speech on both subjective ratings and processing measures. This makes the problem more pressing: even if a system generates sounds natural enough to score well on MOS, systematic phonological errors may go entirely undetected.
%perceived naturalness does not guarantee that a system reproduces language-specific phonotactics, especially in languages where morphophonological alternations shape segment realisation and distribution. 

The dominant evaluation paradigm relies on human listening tests, most commonly MOS and MUSHRA-style protocols \cite{mushra,itu_p800,itu_p910}, as established through benchmarks such as the Blizzard Challenge \cite{blizzard2016}. These protocols provide global ratings rather than targeted evidence about whether a system preserves specific sound contrasts or grammatical alternations. Decontextualized listening tests assume a stable gold standard for speech quality that does not exist, and evaluation should instead be task-specific and diagnostic \cite{wagner2019_TTSevaluation}. \cite{maguer2024moslimits}'s review of MOS studies reveals further limitations, including listener variability, scale anchoring effects, and the collapse of multidimensional speech quality into a single uninformative scalar. \cite{kirkland2023stuck} surveyed 133 recent TTS papers and identified that most do not report basic methodological details such as scale labels or increments, and showed experimentally that even small changes in test instructions can significantly alter scores. 
%Neural TTS systems, including Tacotron 2 \cite{shen2018tacotron2}, FastSpeech \cite{ren2019fastspeech}, and VITS \cite{kim2021vits} have been evaluated primarily using MOS. 

Automatic alternatives include mel cepstral distortion, spectral/F0 error, perceptual metrics such as PESQ \cite{pesq_itu_p862} and STOI \cite{taal2011stoi}, neural MOS predictors \cite{mosnet2019,nisqa2021,utmos2022,cooper2022generalization}, and ASR-based intelligibility (WER/CER). Speaker similarity metrics based on embedding comparison \cite{jia2018transfer} extend evaluation to voice identity but do not address phonological structure. Work on Indian language TTS, including the IndicTTS corpus \cite{kumar2023_indictts}, and multilingual benchmarks such as Common Voice \cite{ardila-etal-2020-commonvoice} have relied on MOS and WER throughout. While useful for tracking overall quality and consistency, none of these measures tests whether a system respects the phonological patterns that distinguish languages, varieties, or accents. Segment-level intelligibility methods such as the Modified Rhyme Test \cite{house1965mrt} and the Diagnostic Rhyme Test \cite{voiers1983drt} established an early precedent for targeted phonological evaluation \cite{nye1973mrt_tts}. Phonological features have been used as training representations in low-resource ASR to support cross-lingual transfer \cite{dalmia_2020}, and probing classifiers have analyzed what phonological information neural speech models encode \cite{belinkov2019analysis,deng1997universalphonfeat,probing_phonology}, but systematic phonological diagnostic testing for multilingual TTS remains rare. 

We address this gap using Assamese Advanced Tongue Root (ATR) vowel harmony as a test case. In Assamese, [+ATR] suffix vowels (/i,u/) trigger harmony on [-ATR] stem vowels (/\textipa{\textepsilon,\textopeno,\textupsilon}/), constraining which vowel qualities can co-occur within a word \cite{mahanta_directionality_2008}. Vowel harmony is part of the language's grammar, and a TTS voice may score well on standard metrics while neutralizing or misplacing harmony-conditioned contrasts. Following cross-linguistic findings on ATR acoustics \cite{hess1992assimilatory,olejarczuk2019acoustic}, [+ATR] vowels are produced with a lower first formant frequency (F1) and narrower first formant bandwidth (B1) than their [-ATR] counterparts. A TTS system that fails to maintain these distinctions will produce measurable F1 shifts that a classifier trained on human speech can detect, even when the output sounds acceptable to a listener. Such failures matter for authenticity, educational uses, and inclusive speech technologies that aim to represent linguistic diversity rather than merely approximating it. We evaluate Meta's Massively Multilingual Speech (MMS) TTS \cite{pratap2024_metamms}, a VITS-based system covering 1100+ languages, using a fixed random seed to ensure deterministic outputs across runs. We present a two-task evaluation pipeline that (i) learns an acoustic-to-phonology mapping from human speech, (ii) applies it cross-domain to synthesized speech, and (iii) summarizes directional mismatches at the vowel and word levels. The methodology is intended to be replicable and adaptable to other phonological systems with measurable acoustic correlates. \footnote{Source code and sample dataset:
\url{https://github.com/snehagitrep/TTSEvalVH_interspeech2026.git}}

\section{Materials \& Methods}
\subsection{Data and model setup}

\textbf{Human benchmark.} We created the human benchmark corpus by recording the speech of $14$ adult native Assamese speakers from the upper Assam region ($8$ females, $6$ males). Each participant read out target words (X) embedded in a carrier frame (\textit{`moi X buli kolu'}, corresponding to English \textit{`I say X'}). We manually sliced the target words from the carrier sentences and segmented the vowels in Praat \cite{BoersmaWeenink2026_praat}. FormantPro \cite{xu2018formantpro} was used to extract the first three formant measurements (F1, F2, F3), the first formant bandwidth (B1) at the 50\% temporal midpoint of each vowel, along with total vowel duration (ms). We applied speaker-level Lobanov-normalization to the formant measures to remove between-speaker physiological variation before cross-speaker modeling. Binary ATR labels ([+ATR], [-ATR]) and two additional features were assigned to each word token following the phonetic and phonological descriptions of Assamese vowels \cite{mahanta_directionality_2008}: vowel height (ordinal: high=0, mid=1, low=2) and backness (binary: front=0, back=1). Tokens were excluded if they fell outside the defined outlier bounds \footnote{F1: $150–1200$ Hz; F2: $500–3500$ Hz; F3: $1500–4500$ Hz; B1 $\leq$ $400$ Hz}, resulting in $8125$ vowel tokens ($4793$ [+ATR], $3332$ [-ATR]) for Task 1.

\begin{table}[th]
\caption{Example stimuli with word tokens and assigned harmony categories.}
 \label{tab:example}
 \centering
   \resizebox{8cm}{!}{  
 \begin{tabular}{llcccc}
    \toprule
\textbf{Stem} & \textbf{Gloss} &\textbf{Suffix} & \textbf{Harmonized}&\textbf{Gloss} & \textbf{Category} \\   
\midrule
    d\textepsilon k\textsuperscript{h} & `to see' & -i & dek\textsuperscript{h}i & `I-see' & AgrYesMixNo\\
    %dil\textepsilon & `' & -t\textupsilon & dil\textepsilon t\textupsilon & `' & AgrNoMixYes \\
    z\textupsilon n\textscripta k & `firefly-M' & -i &  z\textupsilon n\textscripta ki & `firefly-F' & AgrYesMixYes\\
    rup & `silver' & -\textscripta li & rup\textscripta li & `silvery' & AgrNoMixYes\\
    \bottomrule
  \end{tabular}}
\end{table}

\textbf{TTS dataset.} Target words in the same carrier frame were synthesized using Meta's MMS TTS model for Assamese (mms-tts-asm) via Hugging Face \footnote{source code available at \href{https://huggingface.co/facebook/mms-tts-asm}{huggingface.co/facebook/mms-tts-asm}.} with a fixed random seed to ensure deterministic outputs across runs. It is one of the few openly available systems with Assamese support, making the evaluation reproducible. Outputs were synthesized at $16$ kHz mono. The TTS dataset consists of $114$ words, of which $80$ overlap with the human dataset and $34$ are unique. The unique words include harmonizing and non-harmonizing minimal pairs for testing the classifier's prediction accuracy on unseen lexical items. We followed the same annotation and acoustic measurement protocol as for human speech. Since MMS TTS produces a single consistent voice, \textit{z}-normalization was applied globally across the complete TTS corpus, using the corpus-wide mean and standard deviation for each formant (i.e., \textit{z}-scoring within a single speaker’s vowel space). Tokens falling outside the same outlier bounds were excluded from the analysis ($281$ vowel tokens in total; $199$ [+ATR] and $82$ [-ATR]).

%\begin{figure*}
   % \centering
   % \input{plot_spectogram}
  %  \caption{Spectrograms (0–1500 Hz) of [leteku] (AgrYesMixNo) produced by a human speaker (left) and MMS TTS (right), with F1 tracks (yellow) and the approximate +ATR/–ATR boundary at 525 Hz (dashed white).}
   % \label{fig:spectrogram}
%\end{figure*}

\textbf{Stimulus design.} We assigned three harmony categories to the target words based on the surface ATR agreement pattern and the presence of underlying ATR mixing: (i) \textit{AgrYesMixNo} (all vowels agree in ATR class; no mixing) (N=$2102$ in the human dataset and $81$ in the TTS data) and, (ii) \textit{AgrYesMixYes} (surface agreement but mixed ATR values present) (N=$485$ in humans and $25$ in TTS) and (iii) \textit{AgrNoMixYes} (mixedness present and ATR disagreement)(N=$382$ in humans and $8$ in TTS) (see Table \ref{tab:example} for the three harmony categories with example stems and suffixed forms). These word-level labels are used only for Task 2 and for stratified breakdowns of the faithfulness audit. 
%Each target word was embedded in a carrier sentence and synthesised in three variants: original carrier, apostrophe-normalized carrier, and isolated target word. We analysed only the original carrier variant after extracting the target words using Praat, as it provides the most naturalistic phonological context. 

\subsection{Task 1: vowel-level ATR classification}
The primary goal of the first classification task (Task 1) is to characterize whether the acoustic-to-ATR mapping learned from human speech transfers to synthesized speech without significant loss in classification accuracy. We trained a logistic regression (LR; L2 penalty, C=1.0, class-weight=balanced, LBFGS solver) and a random forest classifier (RF; 200 estimators, min$\_$samples$\_$leaf=$5$, class-weight = balanced) on $7$ features (first three normalized formants, B1, duration, height, backness) to predict binary ATR class from acoustic features and then apply it cross-domain to synthesized speech. Four cross-domain transfer directions are evaluated to separate within-domain performance from domain shift. Human$\rightarrow$Human (H$\rightarrow$H), the within-domain reference, uses 5-fold GroupKFold cross-validation by speaker. Human$\rightarrow$TTS (H$\rightarrow$TTS) trains on all human data and tests on TTS. TTS$\rightarrow$TTS uses 5-fold stratified cross-validation on the TTS corpus alone. TTS$\rightarrow$Human trains on TTS and tests on human speech. The gap between H→H and H$\rightarrow$TTS quantifies how much the acoustic structure of TTS vowels diverges from human norms for the same phonological categories. The gap between TTS$\rightarrow$TTS and TTS$\rightarrow$H provides the same diagnosis from the opposite domain. Performance is reported as accuracy (Acc) and macro-averaged F1 (macro-F1).%; standard deviations across folds are reported for cross-validation conditions.

\subsection{Phonological faithfulness audit}
We define phonological faithfulness as the degree to which a TTS system's acoustic output preserves the category-level contrasts specified by the phonological ground-truth of the input text. The audit compares, for each vowel token (N=$8053$ after omitting misaligned tokens), the gold ATR label from phonological transcription against the ATR label inferred from acoustics by the Task 1 classifier. A mismatch is recorded when these two labels disagree. We distinguish two error directions: \textit{overgeneration} (when the TTS produces a [+ATR] surface for an underlying [-ATR] vowel: gold [-ATR] $\rightarrow$ predicted [+ATR]) and \textit{underproduction} (when the system fails to realize [+ATR] where harmony requires it: gold [+ATR] $\rightarrow$ predicted [-ATR]). An asymmetry between these rates indicates a directional bias in how the synthesized acoustics relate to the intended phonological categories. 
For TTS, predictions are generated by the LR classifier trained on all human tokens. For human speech, predictions are generated out-of-fold using the same 5-fold speaker-disjoint scheme as Task 1, ensuring each human token is predicted by a model that never saw that speaker. This makes the human mismatch rate a genuine reference point for classifier uncertainty under speaker-disjoint evaluation. We report overall mismatch rates by harmony type and by vowel identity.

\subsection{Task 2: word-level harmony classification}

Task 2 classifies each word instance into one of three harmony types based on its vowel sequence. Word-level features are computed by aggregating vowel-level measurements within each utterance. Set A (acoustic aggregates; 11 features) summarizes the word’s acoustics using the mean and standard deviation of the normalized formant features, B1, and duration across vowels, along with F1 of the first and last vowel and the vowel count. Set B (ATR sequence summary; 7 features) captures the ATR profile of the vowel sequence: counts of [+ATR] and [-ATR] vowels, proportion [+ATR], binary entropy, majority ATR, an all-agree flag, and the number of ATR switches across the ordered vowel sequence. Set B is computed in two versions: B$\_$gold uses ground-truth ATR labels from phonological transcription, while B$\_$pred uses ATR labels predicted by the Task 1 classifier. For human data, B$\_$pred is derived from speaker-disjoint out-of-fold predictions to avoid leakage; for TTS, B$\_$pred is generated by the LR classifier trained on all human tokens.
We report A, B$\_$pred, A+B$\_$gold, and A+B$\_$pred. Human evaluation uses 5-fold speaker-disjoint CV ($2,969$ instances); TTS evaluation uses Human$\rightarrow$TTS transfer ($114$ instances), noting that AgrNoMixYes is much less frequent (N=$8$). Results are reported for the RF classifier,  which handles the three-class task and nonlinear feature interactions better than LR.

\section{Results}

\subsection{Task 1: vowel-level ATR classification}

Table \ref{tab:task1_crossdomain} reports the accuracy (Acc) and macro-F1 for both models across the four directions. 

\begin{table}[th]
  \caption{Cross-domain ATR classification with Lobanov-normalized acoustic features.}
  \label{tab:task1_crossdomain}
  \centering
    \resizebox{8cm}{!}{  
  \begin{tabular}{llrrrr}
    \toprule
\textbf{Model} & \textbf{Metric} & \textbf{H$\to$H} & \textbf{H$\to$TTS} & \textbf{TTS$\to$TTS} & \textbf{TTS$\to$H} \\   \midrule
    \multirow{2}{*}{LR} & Acc & 81.7\% & 83\% & 86.5\% & 79.8\% \\
                        & macro-F1  & 0.81 & 0.81 & 0.84 & 0.77 \\
    \addlinespace
    \multirow{2}{*}{RF} & Acc & 90.5\% & 74.7\% & 87.5\% & 80.8\% \\
                        & macro-F1  & 0.90 & 0.73 & 0.85 & 0.78 \\
    \bottomrule
  \end{tabular}}
\end{table}

We observe two different outcomes from the cross-domain transfer using the linear (LR) and the non-linear (RF) classifiers. For LR, H$\to$H and H$\to$TTS are nearly identical (Acc $\simeq$  82\% in both conditions and macro-F1 is flat at 0.81). These maintained scores mean the acoustic-to-ATR mapping is stable across domains, but this does not imply phonological faithfulness. RF achieves higher within-domain performance (90.5\% Acc) but shows a considerably larger transfer gap (74.7\%  in H$\to$TTS). This observation indicates that the model may be learning more complex, speaker-conditioned decision boundaries that do not generalize well to a single synthetic voice. 

TTS-only cross-validation (TTS$\to$TTS) has a macro-F1 of 0.842 (LR) and 0.847 (RF) , broadly comparable to the human CV scores. Given that the two datasets differ in size, speaker count, and label balance, we do not attribute this similarity to the TTS vowel system's properties. The TTS$\to$H direction performs with above-chance accuracy (Acc $\simeq$ 80\% -- 81\%) scores, though the per-vowel breakdown reveals that /\textupsilon/ is particularly poorly recovered in this direction (Acc 0.9\% for RF, 6.3\% for LR). As /\textupsilon/ has very few tokens in the TTS output (N=7), this result reflects data sparsity rather than a reliable estimate of classifier behavior for that category. At the individual vowel-level, we see that /i/ and /\textscripta/ vowels are classified near-perfectly (98\% and 100\%, respectively, for LR) in the H$\to$TTS domain. This pattern is consistent with [+high] and [-high] vowels occupying distinct regions of the formant space. The mid [+ATR] vowels /e/ and /o/ are considerably harder (Acc 69\% and 66\% , respectively), a pattern that anticipates the faithfulness audit findings discussed in the following section. /u/ is among the harder vowels (66.2\%), but in H$\to$TTS it reaches 80\% accuracy, suggesting the TTS generates this vowel with more consistent acoustic properties than natural speech across speakers.
%/u/ classifies more accurately in H$\to$TTS (80\%) than in human CV accuracy for that vowel. %We use LR for the faithfulness audit because its cross-domain transfer is stable and interpretable.
%Lexical overlap analysis shows that Human$\to$TTS performance is higher on overlapping words than on unique words for all model/feature configurations (e.g., LR (F1): 0.84 overlap vs 0.76 unique; RF (F1): 0.81 overlap vs 0.63 unique). This indicates that transfer to unseen lexical items is harder, while remaining viable, particularly for LR with normalized features. 

\subsection{Phonological faithfulness audit}

In Table \ref{tab:ruleaudit}, we report the overall mismatch rate (proportions of tokens where the ground-truth labels do not match the predicted labels), with error directions broken down into overgeneration (when ground-truth [-ATR] label is predicted as [+ATR]), and underproduction (when ground-truth [+ATR] is predicted as [-ATR]).

\begin{table}[th]
  \caption{Phonological faithfulness audit: overall mismatch rates}
  \label{tab:ruleaudit}
  \centering
  \resizebox{8cm}{!}{
  \begin{tabular}{lrrrr}
    \toprule
    \textbf{Domain}& \textbf{N} & \textbf{Mismatch} &\textbf{Overgeneration} & \textbf{Underproduction}\\
        & & &   $(-\to+)$ & $(+\to-)$ \\
    \midrule
    Human (OOF) & 8053 & 0.18 & 0.09 & 0.09 \\
    TTS         & 281  & 0.16 & 0.02 & 0.14 \\
    \bottomrule
  \end{tabular}}
\end{table}

\begin{figure}[th]
    \centering

     \begin{subfigure}[b]{\columnwidth}
        \centering
        \includegraphics[width=\columnwidth]{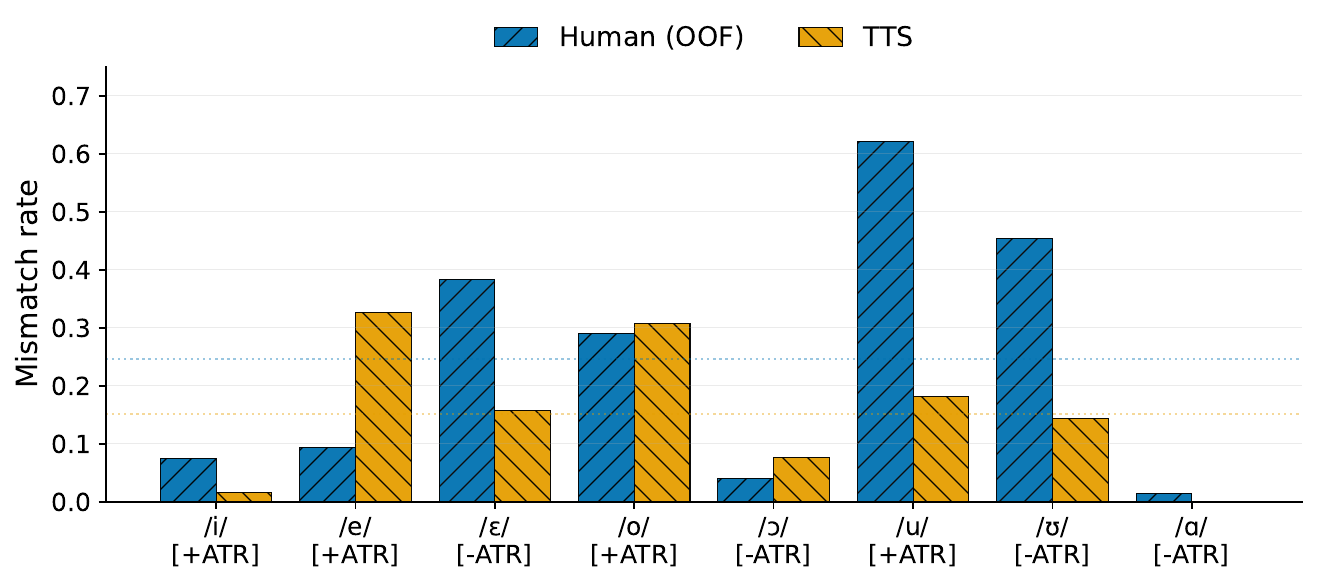}
        \caption{Per-vowel mismatch rate.}
        \label{fig:pervowelmismatchaudit}
    \end{subfigure}
    \vspace{0.4em}

    \begin{subfigure}[b]{\columnwidth}
        \centering
        \includegraphics[width=\columnwidth]{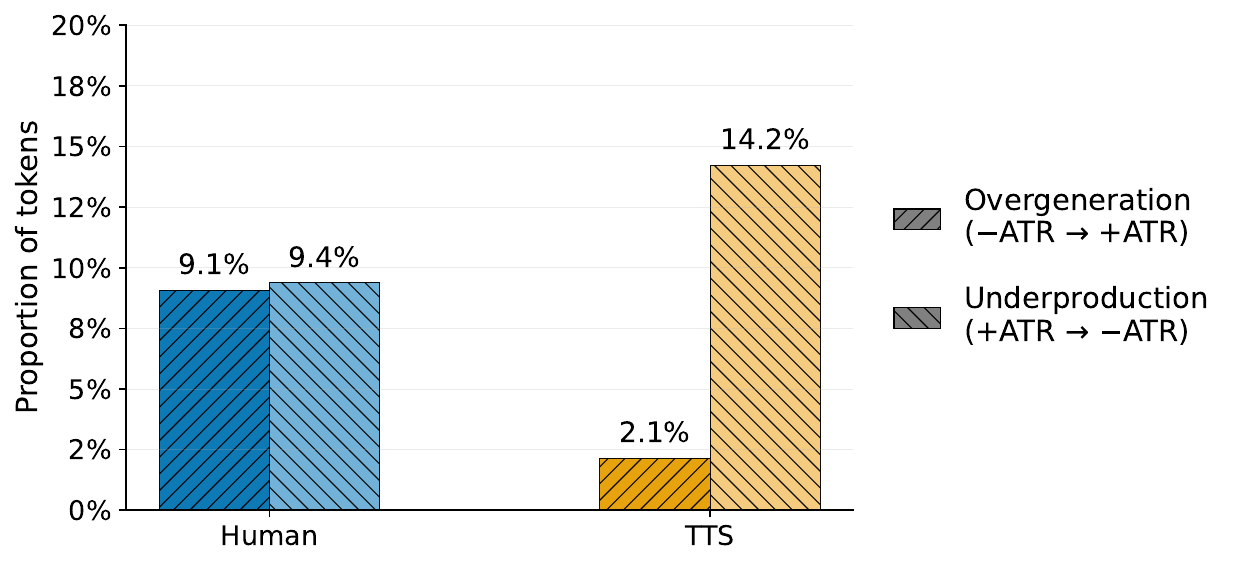}
        \caption{Aggregate error directionality.}
        \label{fig:faithfulnessauditdirection}
    \end{subfigure}
   
    \caption{Faithfulness audit error directionality. (a) Mismatch concentrates in mid [+ATR] vowels /e/ and /o/ in TTS, whereas human mismatch is highest for /u/, /\textupsilon/, and /\textepsilon/. (b) Human errors are roughly symmetric; TTS errors show a 7:1 underproduction-to-overgeneration ratio . }
    \label{fig:auditcombined}
\end{figure}

Our out-of-fold human benchmark shows a mismatch rate of $0.185$ (N=$8053$) and nearly symmetric error directions, with overgeneration at $0.091$, and underproduction at $0.094$.  This symmetry is expected, given the acoustic overlap between ATR categories in mid vowels, and it establishes the benchmark for classifier uncertainty under speaker-disjoint evaluation.  The TTS domain, by contrast, is quite different \footnote{Of  $199$ gold [+ATR] tokens, $40$ ($20.1\%$) were predicted [-ATR]; of $82$ gold [-ATR] tokens, $6$ ($7.3\%$) were predicted [+ATR].}. The overall mismatch rate is slightly lower than in humans ($0.164$; N = $281$), but the errors are one-sided. The tokens gold-labeled (ground-truth) as [+ATR] are frequently predicted as [-ATR] ($0.142$; $40/199$), while the frequency of the gold [-ATR] predicted as [+ATR] is much lower ($0.021$; $6/82$) (Figure \ref{fig:faithfulnessauditdirection}).  This $7:1$ asymmetry has no counterpart in human speech. The lower overall mismatch in TTS reflects that most TTS tokens are gold [+ATR] and that the classifier mostly gets them right when they are acoustically unambiguous. We treat the directional imbalance as the diagnostic signal, not the overall rate. The difference in error directionality between domains is statistically significant ($\chi^{2}(1) = 21.94,\; p < 0.001$). Human errors are split almost evenly, but almost 87\% of TTS errors are due to underproduction. 

%The directional distribution of errors differs significantly between domains ($\chi\textsuperscript{2}$(1) = $21.94, p < 0.001$): in human speech, mismatches split nearly evenly between overgeneration and underproduction, whereas in TTS, underproduction accounts for 87\% of all errors. 

%We argue that the lower overall mismatch in TTS does not indicate better phonological faithfulness, but reflects the fact that most TTS tokens happen to be gold [+ATR] and the classifier mostly gets those right when they are acoustically unambiguous.

\begin{figure}[!ht]
\centering
\begin{minipage}[t]{0.48\columnwidth}
  \centering
  \includegraphics[width=\linewidth]{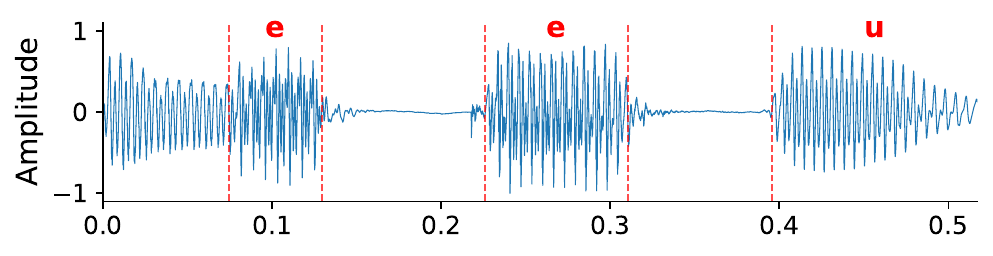}\\[1pt]
  \includegraphics[width=\linewidth]{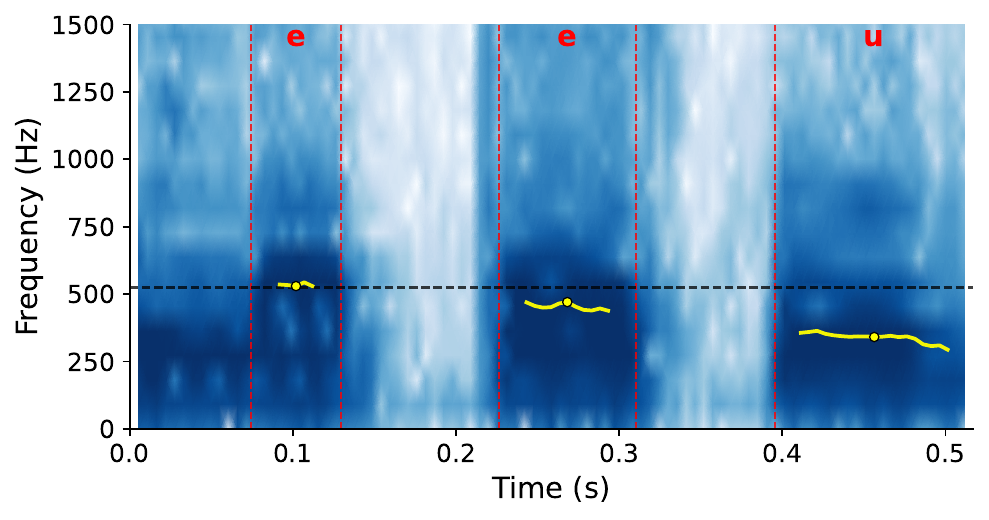}\\[2pt]
  {\footnotesize (a) [leteku] (human speech)}
\end{minipage}%
\hfill
\begin{minipage}[t]{0.48\columnwidth}
  \centering
  \includegraphics[width=\linewidth]{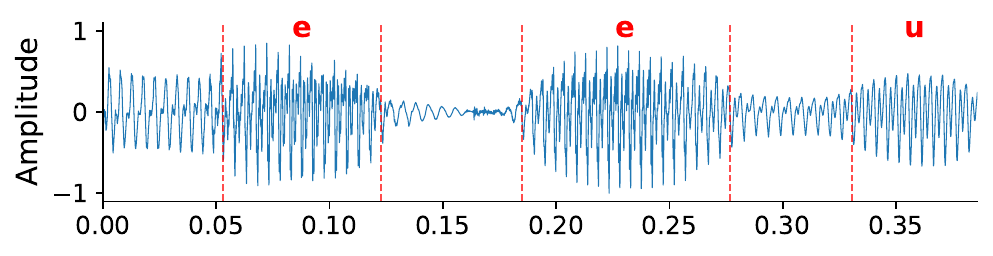}\\[1pt]
  \includegraphics[width=\linewidth]{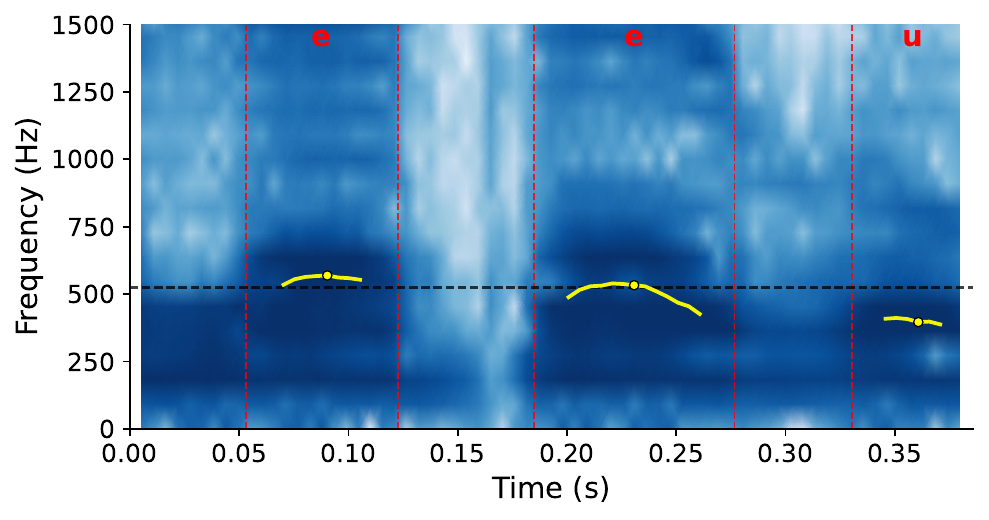}\\[2pt]
  {\footnotesize (b) [leteku] (TTS speech)}
\end{minipage}
\caption{Spectrograms ($0-1500$ Hz) of [leteku] (AgrYesMixNo) produced by a human speaker (left) and MMS TTS (right), with F1 tracks (yellow) and the approximate [+ATR]/[-ATR] boundary derived from the human benchmark (dashed white). In TTS, the F1 of both /e/ tokens sits near or above the boundary, consistent with the underproduction pattern reported in Table~\ref{tab:ruleaudit}.}
\label{fig:spectrogram}
\end{figure}

The vowel-level observation clarifies where the underproduction is concentrated (Figure \ref{fig:pervowelmismatchaudit}). In TTS, /i/ (mismatch: $0.016$) and /\textscripta/ ($0.000$) are correctly classified, consistent with the cross-domain task 1 results for high and low vowels. However, the mid [+ATR] vowels /e/ and /o/ show the highest mismatch rates ($0.327$; N=$52$ and $0.308$; N=$52$, respectively). This means roughly one in three tokens of each vowel is predicted as [-ATR] by a classifier trained on human speech. Among the [-ATR] vowels in TTS, mismatch is low for /\textopeno/ ($0.077$) and moderate for /\textepsilon/ ($0.158$). Earlier, we noted that /u/ ($0.662$), /\textupsilon/ ($0.453$), and /\textepsilon/ ($0.383$) are the hardest vowels for the classifier in the human data. The contrastive pattern of TTS suggests that the increased mismatch on /e/ and /o/ is not just a reflection of those vowels being inherently difficult to classify. The mismatch between [+ATR] mid-vowels in TTS indicates that their acoustic realizations fall within the [-ATR] region of the human-trained classifier's decision boundary, despite their [+ATR] specification. Figure \ref{fig:spectrogram} illustrates the pattern for [leteku] where both /e/ tokens in the TTS output sit near or above the [$\pm$ ATR] F1 boundary derived from human speech.

%Overall, we observe a pattern: when the token is labeled as [+ATR], the acoustics are often not confidently [+ATR] according to the human-trained classifier, with the largest concentration of mismatches in the mid-vowel region (/e,o/) (see Figure \ref{fig:spectrogram} for example illustration). %This pattern is consistent with observations during manual segmentation, where mid [+ATR] vowels (/e,o/) appeared to dominate the TTS outputs even in harmony contexts that would license /\textepsilon/ or /\textopeno/ (Figure \ref{fig:spectrogram}). 

\subsection{Task 2: word-level harmony classification}

The second classification task tests whether the token-level patterns identified in the faithfulness audit are also retained in the word-level harmony classification. For human data, B$\_$pred uses speaker-disjoint out-of-fold predictions, and for TTS, it uses ATR predictions from the LR classifier trained on all human tokens in task 1. If TTS realized intended ATR categories faithfully, A+B$\_$gold and A+B$\_$pred should perform similarly. Table \ref{tab:task2_crossdomain} reports the RF classifier's performance on task 2.

%We aggregate vowel-level measurements into word-level feature sets: Set A (11 acoustic aggregates: mean/SD of formants, duration, vowel count, position effects) and Set B (7 ATR distribution features: counts, proportions, entropy, switches). Set B is computed in two steps: B$\_$gold from phonological transcription and B$\_$pred from Task 1 classifier predictions. 

\begin{table}[th]
  \caption{Word-level harmony classification. Acc / macro-F1 reported for 
  each transfer direction. \textcolor{red!60!black}{$\downarrow$} marks the 
  drop below the acoustic-only baseline on H$\to$TTS transfer.}
  \label{tab:task2_crossdomain}
  \centering
  \resizebox{\columnwidth}{!}{
  \begin{tabular}{p{3.9cm}lrr}
    \toprule
    \textbf{Feature description} & \textbf{Set} & \textbf{H$\to$H} & \textbf{H$\to$TTS} \\
     %& & Acc (macro-F1) & Acc (macro-F1) \\
    \midrule
    Acoustic features only (F1, F2, B1, duration) & A & 84.0\% (0.76) & 71.1\% (0.64) \\
    \addlinespace
    Phonological ATR sequence summary, predicted labels (proportion +ATR, entropy, switches) & B$\_$pred & 62.8\% (0.54) &\textbf{ 54.4\% (0.50)} \textcolor{red!60!black}{$\downarrow$}\\
    \addlinespace
    Acoustic + gold ATR sequence & A+B$\_$gold & 88.8\% (0.83) & \textbf{58.8\% (0.49)} \textcolor{red!60!black}{$\downarrow$} \\
    \addlinespace
    Acoustic + predicted ATR sequence & A+B$\_$pred & 84.2\% (0.77) & 69.3\% (0.62) \\
    \bottomrule
  \end{tabular}}
\end{table}

A+B$\_$gold achieves the best within-domain performance ($0.84$), which is expected since gold ATR labels directly encode the harmony structure. On transfer to TTS, however, the A+B$\_$gold drops to $0.49$, below the acoustic-only baseline of $0.64$. A+B$\_$pred shows the opposite pattern. Its performance drops within-domain ($0.77$ macro-F1) but holds up better on TTS ($0.62$). The transfer gap in macro-F1 decreases from $0.34$ (A+B$\_$gold) to $0.15$ (A+B$\_$pred). 

What the gap between A+B$\_$gold and A+B$\_$pred in H$\to$TTS tells us is methodologically useful. Gold ATR labels encode what a word's vowels are supposed to sound like phonologically. When those labels are used to train a word-level classifier on human speech and then applied to TTS, the classifier encounters TTS words whose acoustic ATR profiles, as captured by B$\_$pred,  do not consistently match what gold labels would predict. The gap (+0.13 macro-F1 in favor of A+B$\_$pred) is a quantitative measure of that misalignment. A TTS system that faithfully realized its intended ATR categories would show a much smaller difference between these two conditions, because the acoustic predictions would closely track the gold labels. The fact that A+B$\_$pred outperforms A+B$\_$gold on TTS is consistent with the interpretation that the acoustic ATR profile diverges from the intended phonological structure, with Set A's acoustic aggregates contributing most of the signal.
%the acoustic ATR profile of the synthesized output diverges from the phonological structure the labels represent. The acoustic aggregates in Set A carry substantial information about harmony type on their own, which is why the A-only baseline is competitive and why A+B$\_$pred gains most of its advantage from combining both sources rather than from the phonological features alone.

\section{Conclusion}

Our results show that MMS TTS underproduces the acoustic correlates of [+ATR] in mid vowels, with a 7:1 underproduction-to-overgeneration ratio absent from human speech. This bias persists at the word-level: the drop in accuracy and macro-F1 from A+B$\_$gold relative to A+B$\_$pred on TTS transfer (Table \ref{tab:task2_crossdomain}) reflects that the phonological categories the system is supposed to realize do not consistently align with the acoustic output. For languages in which grammatical context conditions segment quality, a phonologically informed evaluation layer may provide useful complementary information. The audit format we propose is one possible way to give system developers more targeted feedback about where and how a system deviates. This study evaluates a single TTS system for a single phonological phenomenon in a single language using a small, class-imbalanced TTS dataset. These factors constrain the specific error rates and the strength of the claims we can make, but not the general approach, which in principle extends to any phonological contrast with measurable acoustic correlates and a human benchmark.

%We note that evaluating a single TTS system on a single phonological phenomenon in a single language poses a limitation on generalizability. Moreover, the TTS dataset is small and class-imbalanced, limiting the generalizability of the specific error rates reported here; however, they do not undermine the methodology, which is our primary contribution. Our pipeline generalizes in principle to any phonological contrast with measurable acoustic correlates and a human speech baseline. 

\section{Use of Generative AI Disclosure}
Generative AI was used to assist with code auto-completion, minor text editing, and grammar polishing. However, all scientific content, code implementations, results, and analyses were proposed, verified, and finalized by the authors.

\bibliographystyle{IEEEtran}
\bibliography{mybib}

\end{document}